\documentclass[lettersize,journal]{IEEEtran}
\usepackage{amsmath,amsfonts}
\usepackage{algorithmic}
\usepackage{algorithm}
\usepackage{array}
\usepackage[caption=false,font=normalsize,labelfont=sf,textfont=sf]{subfig}
\usepackage{textcomp}
\usepackage{stfloats}
\usepackage{url}
\usepackage{verbatim}
\usepackage{graphicx}
\usepackage{multirow}
\usepackage{colortbl}
\usepackage[table,xcdraw]{xcolor}
\usepackage{threeparttable}
\usepackage{hhline}
\usepackage{threeparttable}
\usepackage{array}  

\begin{document}

\title{Modality Unified Attack for Omni-Modality Person Re-Identification}

\author{Yuan Bian, Min Liu, Yunqi Yi, Xueping Wang, Yunfeng Ma and Yaonan Wang
  \thanks{This work was supported in part by the National Natural Science Foundation of China under Grant 62425305, U22B2050 and 62221002, in part by the Science and Technology Innovation Program of Hunan Province under Grant 2023RC1048, in part by the Hunan Provincial Natural Science Foundation of China under Grant 2024JJ3013. (\textit{Corresponding author: Min Liu}) }
  \thanks{
    Yuan Bian, Min Liu, Yunqi Yi, Yunfeng Ma, and Yaonan Wang are with the College of Electrical and Information Engineering at Hunan University and National Engineering Research Center of Robot Visual Perception and Control Technology, Changsha, Hunan, China. E-mail: yuanbian$@$hnu.edu.cn; liu\_min$@$hnu.edu.cn; y0512321$@$hnu.edu.cn; ismyf$@$hnu.edu.cn; yaonan$@$hnu.edu.cn.

    Xueping Wang is with the College of Information Science and Engineering at Hunan Normal University, and Hunan Provincial Key Laboratory of Intelligent Computing and Language Information Processing, Changsha, Hunan, China. E-mail: wang\_xueping$@$hnu.edu.cn.
    
  }
}



\maketitle

\begin{abstract}
  Deep learning based person re-identification (re-id) models have been widely employed in surveillance systems. Recent studies have demonstrated that black-box single-modality and cross-modality re-id models are vulnerable to adversarial examples (AEs), leaving the robustness of multi-modality re-id models unexplored. Due to the lack of knowledge about the specific type of model deployed in the target black-box surveillance system, we aim to generate modality unified AEs for omni-modality (single-, cross- and multi-modality) re-id models. Specifically, we propose a novel Modality Unified Attack method to train modality-specific adversarial generators to generate AEs that effectively attack different omni-modality models. A multi-modality model is adopted as the surrogate model, wherein the features of each modality are perturbed by metric disruption loss before fusion. To collapse the common features of omni-modality models, Cross Modality Simulated Disruption approach is introduced to mimic the cross-modality feature embeddings by intentionally feeding images to non-corresponding modality-specific subnetworks of the surrogate model. Moreover, Multi Modality Collaborative Disruption strategy is devised to facilitate the attacker to comprehensively corrupt the informative content of person images by leveraging a multi modality feature collaborative metric disruption loss. Extensive experiments show that our MUA method can effectively attack the omni-modality re-id models, achieving 55.9\%, 24.4\%, 49.0\% and 62.7\% mean mAP Drop Rate, respectively.
\end{abstract}

\begin{IEEEkeywords}
Modality unified attack, Omni-Modality re-id, Multi Modality Collaborative Disruption.
\end{IEEEkeywords}

\section{Introduction}
\IEEEPARstart{P}{erson} re-identification (re-id), which seeks to retrieve a specific individual in video surveillance, has drawn significant attention in recent years \cite{ye2021deep,zheng2016person}. Over the past decades, numerous \textbf{single-modality re-id} methods have been proposed for RGB image matching \cite{10262018,10130650,9693924,zheng2015partial,liu2023weakly,sun2018beyond,tang2024ped, wang2019spatial,li2021diverse,he2021transreid,bian2023occlusion,liu2024two}, showcasing promising results. However, these RGB-based methods often struggle in low-light or dark environments. To overcome these limitations, the \textbf{cross-modality re-id} models have been introduced \cite{ye2020visible,wu2017rgb,9963944,wang2020cross,li2020infrared,yang2023towards}, which aim retrieving the same person across RGB and near-infrared (NI) modality images. More recently, researchers have been developing \textbf{multi-modality re-id} methods to leverage complementary information from RGB, NI, and thermal infrared (TI) images to create even more robust and effective re-id systems \cite{zheng2021robust,wang2022interact,zhang2024magic,wang2024top}.

\begin{figure}[t]
  \centering
  \includegraphics[width=1.0\columnwidth]{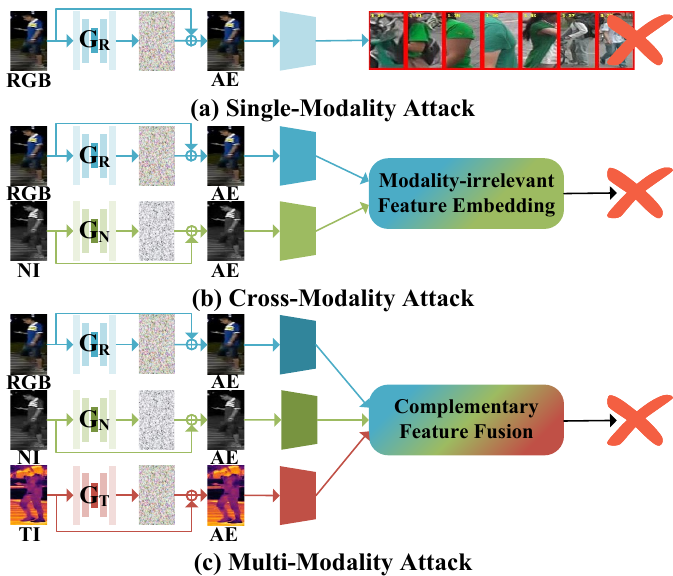}
  \caption{The diagram of modality unified adversarial attack for omni-modality (single-, cross- and multi-modality) person re-id models. $G_{R}$, $G_{N}$ and $G_{T}$ are modality unified adversarial generators for RGB, NI and TI images, which can generate AEs to successfully attack all three types of re-id models, e.g., the RGB modality unified adversarial generator $G_{R}$ can generate RGB AEs that simultaneously mislead single-, cross- and multi-modality re-id models.}
  \label{fig:1}
\end{figure}

Notably, almost all single-modality, cross-modality, and multi-modality person re-id models are deep learning based, which unfortunately inherit the vulnerabilities of deep neural networks to adversarial samples \cite{goodfellow2014explaining,szegedy2014intriguing, madry2018towards}. This poses a significant threat to public safety, as these systems can be potentially compromised. Consequently, it is essential to investigate the vulnerability of deep learning based re-id models to adversarial samples to ensure the security and reliability of surveillance systems.

Recently, researchers have begun to explore the robustness of re-id methods, with several studies revealing that re-id models are vulnerable to AEs. Specifically, some researchers \cite{zheng2023u,bai2020adversarial,bouniot2020vulnerability,wang2019advpattern,gong2024cross,wang2024decoupled} have demonstrated the susceptibility of re-id models to white-box adversarial metric attacks. Additionally, other researchers have focused on developing transferable AEs against black-box re-id models \cite{yang2022towards,yang2021learning,ding2021beyond,wang2020transferable,subramanyam2023meta,bian2024learning}, taking into account the transferability of attacks across different datasets and models.
Nevertheless, most existing studies have primarily focused on adversarial attacks for single-modality re-id models, with perturbations crafted specifically for RGB images, while research on adversarial vulnerabilities in cross-modality re-id models is limited, and multi-modality re-id models remain largely unexplored.


To comprehensively investigate the robustness of single-modality, cross-modality, and multi-modality person re-id models against malicious attacks, we undertake this study for countering attacks on these three types of re-id models. Furthermore, considering the real-world scenario, an attacker cannot distinguish the type of model applied in black-box re-id systems, but can only recognize the modality of the input image, so we need to train modality unified adversarial attackers to effectively attack any type of re-id models. The modality unified adversarial attacks for omni-modality re-id models are demonstrated in Fig. \ref{fig:1}, where the modality unified adversarial generators can generate AEs for specific modality images that can effectively mislead different categories of re-id models.


For our purpose, we first select a multi-modality re-id model as the surrogate model. This is because multi-modality re-id models typically employ modality-specific subnetworks to extract features of three modality images before complementary feature fusion, enabling us to train three modality unified attackers concurrently by compromising individual modality features independently. 
However, there are two challenges hinder us from generating transferable modality unified AEs across three types re-id models: (1) Different types of models extract distinct features for the same modality images, i.e., single-modality models learn discriminative color information of person clothes, whereas cross-modality models do not, as they require modality-invariant features with NI images to accomplish cross-modality retrieval. Disrupting these distinct features will cause limited transferability. Therefore, \textbf{\emph{how to perturb common features of same modality images for omni-modality models is the first challenge.}}
(2) Multi-modality re-id models integrate features from multiple modality data to improve the robustness of recognition, which compels us to prevent AEs from retaining any compensate information for feature fusion.
Thus, \textbf{\emph{how to maximally destroy useful features of multi modality images for omni-modality models is the second challenge}}.

To address the aforementioned challenges, we propose a novel Modality Unified Attack (MUA) method to achieve effective attack on omni-modality person re-id models. The multi-modality re-id model is selected as the surrogate model to concurrently train three modality unified adversarial generators by disturbing the metric similarity of the intermediate features before feature fusion. Cross Modality Simulated Disruption (CMSD) is introduced to emulate the cross-modality embedding spaces by injecting uncorresponding modality images into modality-specific subnetworks, which prompts attackers to attack common features of the same modality image in omni-modality models. In addition, Multi Modality Collaborative Disruption (MMCD) strategy is presented to guide the generated AEs to contain minimized useful information through the cooperative multi-modality feature constraints. 
Our main contributions are summarized as follows:
\begin{itemize}
    \item To the best of our knowledge, our method is the first work that study the modality unified attack for omni-modality person re-id models.
    \item Cross Modality Simulated Disruption approach is introduced to mimic the feature embeddings across different types of re-id models, facilitating attackers perturbing the shared features of same modality images in omni-modality models.
    \item Multi Modality Collaborative Disruption strategy is devised to steer the generation of AEs that reduce valuable information through customized multi modality feature collaborative metric disruption constraints.
    \item Extensive experiments show that our MUA method can effectively attack the omni-modality re-id models, achieving 55.9\%, 24.4\%, 49.0\% and 62.7\% mean mAP Drop Rate on different omni-modality retrieve settings, respectively.
\end{itemize}




\section{Related Works}

\subsection{Person Re-identification}
Person re-id is first introduced to retrieve specific person from the RGB video systems. Extensive methods have been proposed to extract discriminative features of RGB images in occluded \cite{gao2020pose}, pose variation \cite{liu2018pose}, illumination fluctuation \cite{huang2019illumination} scenarios under strongly supervised \cite{ he2021transreid}, weakly supervised \cite{Wang2021Learning} and unsupervised situations \cite{Wang2021Exploiting}.
To meet the needs of surveillance systems for all-day monitoring, the cross-modality person re-id methods have been studied recently \cite{wu2017rgb,wang2020cross,li2020infrared,yang2023towards}, which mainly use two-stream networks to learn modality-invariant features of RGB and NI images.
Very recently, multi-modality methods \cite{zheng2021robust,wang2022interact,zhang2024magic,wang2024top} have
been developing to enhance the feature robustness in complex visual scenarios by aggregating RGB, NI and TI person images. Multi-modality methods basically utilize three modality-specific feature extractors to derive
modality-specific features, which are then followed by multi-modality fusion operations.

\subsection{Adversarial Attack for Re-id}
The vulnerability of single-modality re-id models has been widely explored, and few researches have focused on the security of cross-modality re-id models. White-box attack for re-id \cite{bai2020adversarial,zheng2023u,bouniot2020vulnerability} were firstly proposed by introducing many metric attack methods. Wang \textit{et al.} \cite{wang2019advpattern} proposed a physical-world attack algorithm for generating adversarial transformable patterns to realize adversary mismatch and target person impersonation attacks. Bai \textit{et al.} \cite{bai2020adversarial} proposed an adversarial metric attack that utilizes probe and gallery feature distances to attack re-id models.
Considering the unknown queries and re-id model in practical, cross-model and cross-dataset transferable adversarial attacks \cite{yang2022towards,ding2021beyond,yang2021learning,wang2020transferable,subramanyam2023meta} are devised.
Wang \textit{et al.} \cite{wang2020transferable} developed a multi-stage discriminator network for cross-dataset general attack learning. Yang \textit{et al.} \cite{yang2022towards} built a combinatorial attack that consists of a functional color attack and universal additive attack to promote the cross-model and cross-dataset transferability of the attack.
These above methods only focus on single-modality person re-id models, only Gong \textit{et al.} \cite{gong2024cross} and Yang \textit{et al.} \cite{yangfeature} introduced universal perturbations for cross-modality person re-id attack. Gong \textit{et al.} leveraged the gradients from diverse modality data to generate universal perturbation for both RGB and NI images, but it is not rational for NI and RGB images share the same universal perturbation, since the imaging principles of these two modality images are different. Yang \textit{et al.} generated universal adversarial perturbations for each modality images, using a Frequency-Spatial Attention Module to emphasize important regional features across domains, and apply an Auxiliary Quadruple Adversarial Loss to increase modality distinction.

\subsection{Multi-modality Attack}
There are few multi-modality attack methods in different fields.
Li \textit{et al.} \cite{li2019cross} presented a cross-modality adversarial example learning algorithm that produces two perturbations for different modalities by decreasing the inter-modality similarity and simultaneously keeping intra-modality similarity. Li \textit{et al.} \cite{li2021adversarial} designed cross-modality triplet construction module to learn cross-modality Hamming retrieval AEs by mining cross-modality positive and negative instances in Hamming space. Wang \textit{et al.} \cite{wang2023targeted} proposed a targeted attack method against deep cross-modality hashing retrieval by seamlessly embedding target semantics and benign semantics into U-Net framework. Shi \textit{et al.} \cite{shi2022multifeature} developed attack for multimodal remote sensing classification networks by training modality-specific GANs. Wei \textit{et al.} \cite{wei2023unified} designed a unified cross-modality attack method that generates adversarial patches to fool visible and infrared object detectors.

Distinct from existing single-modality and cross-modality person re-id attacks and other multi-modality attacks, our method aims to generate modality unified AEs for omni-modality models. To enhance the transferability of AEs across different types of re-id models, Cross Modality Simulated Disruption and Multi Modality Collaborative Disruption methods are innovatively proposed.

\begin{figure*}[t]
    \centering
    \includegraphics[width=2.0\columnwidth]{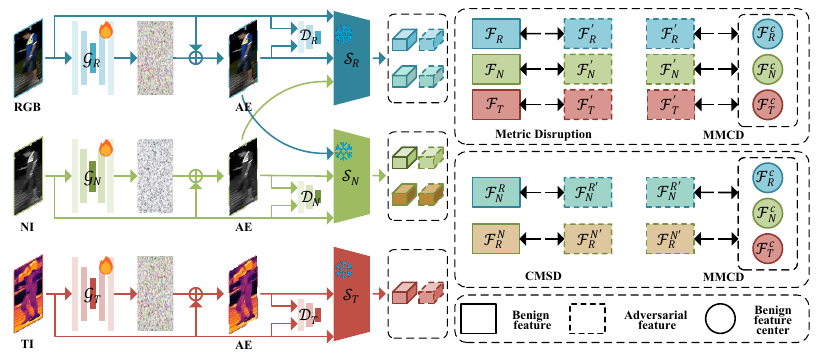}
    \caption{The overall framework of our proposed MUA. The multi-modality model is adopted as surrogate model, and features $\mathcal{F}_h$ extracted by subnetworks $\mathcal{S}_h$ are perturbed before fusion operation. Adversarial attack generator $\mathcal{G}_h$ and discriminator $\mathcal{D}_h$ are trained in GAN scheme to obtain imperceptible and qualify AEs. Metric Disruption, CMSD and MMCD constraints guide the adversarial generator produce transferable AEs across omni-modality re-id models.}
    \label{fig:2}
\end{figure*}

\section{Methodology}
In this section, we first present the problem definition of the modality unified adversarial attack against omni-modality re-id models. The overall framework of our MUA is then introduced and the details of Metric Disruption, Cross Modality Simulated Disruption, Multi Modality Collaborative Disruption are described right after that. Finally, the overall objective function of our method are introduced.

\subsection{Problem Definition}
Person re-id models $\mathcal{M}$ aim to extract discriminative features of query images and gallery images, then the feature similarity scores of them are sorted in decreasing order by sort function $Rank(\cdot)$. The extracted features $\mathcal{M}(x)$ of query images $x$ are supposed to as similar as features $\mathcal{M}(g)$ of target person images $g$ in gallery, within the top-$K$ results by $Rank(\cdot)$.
\begin{equation}
    Rank(\mathcal{M}(x),\mathcal{M}(g))<K.
    \label{eq:eq1}
\end{equation}

The goal of our proposed MUA is to train modality unified adversarial generator $\mathcal{G}_{h}$ for $h$ modality images to successfully mislead the omni-modality re-id models. The adversarial example $x^{\prime}_{h}$ is produced by adding additive perturbation $\mathcal{G}_{h}(x_{h})$ to the query image $x_h$ to attack the re-id models $\mathcal{M}$ for outputting incorrect retrieval images, where the maximum magnitude of perturbations allowed to be added cannot exceed $\epsilon$.
\begin{equation}
    x^{\prime}_{h} = \mathcal{G}_{h}(x_{h})+x_{h},
    \quad\mathrm{s.t.}\|x^{\prime}_{h}-x_{h}\|_\infty\leq\epsilon.
    \label{eq:eq2}
\end{equation}

Specifically, for single-modality re-id model $\mathcal{M}_{S}$, the input query images $x_R$ and target person gallery images $g_R$ are both in RGB modality. Our MUA aims to utilize $G_R$ to attack the single-modality re-id models.
\begin{equation}
    Rank(\mathcal{M}_{S}(x^{\prime}_{R}),\mathcal{M}_{S}(g_{R}))>K,
    \label{eq:eq3}
\end{equation}

For cross-modality re-id model $\mathcal{M}_{C}$, it seeks to retrieve the same person across RGB images $x_R$ and NI images $x_N$. The purpose of our MUA method is to train adversarial generator $G_R$ and $G_N$ to corrupt the performance of visible-to-infrared and infrared-to-visible retrieval settings.
\begin{equation}
    \begin{split}
        Rank(\mathcal{M}_{C}(x^{\prime}_{R}),\mathcal{M}_{C}(g_{N}))>K, \\
        Rank(\mathcal{M}_{C}(x^{\prime}_{N}),\mathcal{M}_{C}(g_{R}))>K.
    \end{split}
    \label{eq:eq4}
\end{equation}

For multi-modality model $\mathcal{M}_{M}$, it takes RGB, NI and TI  images $x_R$, $x_N$, $x_T$ as input to get complementary features to achieve more robust performance. Our MUA adopts adversarial generators $G_R$, $G_N$ and $G_T$ to simultaneously generate AEs for these three modality images to attack multi-modality re-id models.
\begin{equation}
    Rank(\mathcal{M}_{M}(x^{\prime}_{R},x^{\prime}_{N},x^{\prime}_{T}),\mathcal{M}_{M}(g_{R},g_{N},g_{T}))>K.
    \label{eq:eq5}
\end{equation}

\subsection{Overall Framework}
The overall framework is shown in Fig. \ref{fig:2}. To train the adversarial generator for three modalities simultaneously, we utilize a multi-modality model as the agent model. Such agent model contains three modality-specific subnetworks $\mathcal{S}_h$ to extract features $\mathcal{F}_h$ of each modality data $x_{h}$,
\begin{equation}
    \mathcal{F}_{h}=\mathcal{S}_{h}(x_{h}),
    \label{eq:eq6}
\end{equation}
after which a fusion of the three modalities features is performed to achieve the enhanced performance of the re-id. Inspired by some transferable attacks \cite{huang2019enhancing,wang2021feature,zhang2022enhancing} that intermediate feature disruption can enhance the attack performance for black-box models, our MUA intends to destroy these intermediate layer features $\mathcal{F}_h$ before feature fusion to train adversarial generators. Metric Disruption Loss is utlized to perturb the features by pushing the AE feature $\mathcal{F}^\prime_{h}$ away from benign image feature $\mathcal{F}_{h}$. CMSD and MMCD are introduced to enhance the transferability of the generated AEs. The discriminator $\mathcal{D}_h$ for $h$ modality is utilized to support the generation of better quality and imperceptible AEs.

\subsection{Metric Disruption}
Re-id task is a metric learning task, which aims to extract discriminative features of person images and find the target person by similarity metrics. It has no concept of classification decision boundaries in the retrieving process and employs metric functions to measure the pairwise similarity between the probe and gallery images. To corrupt the features of input images effectively, the Metric Disruption method is utilized in our MUA to directly disrupt the similarity of AE features and benign features by pushing them away. Specifically, adversarial Euclidean Distance ($\mathcal{E}$) metric function is adopted, and Metric Disruption Loss is formulated as
\begin{equation}
    \mathcal{L}^h_{MD}=-\mathcal{E}(\mathcal{F}_h,\mathcal{F}^\prime_h).
    \label{eq:eq7}
\end{equation}
\subsection{Cross Modality Simulated Disruption}
The surrogate multi-modality models \cite{wang2024top,zhang2024magic} adopt three independent subnetworks to capture the modality-specific features $\mathcal{F}_h$ of each modality image for complementary feature fusion, so  the feature embedding spaces of features $\mathcal{F}_h$ in Metric Disruption are same with RGB single-modality models. However, it is different from the feature spaces of cross-modality models, since they intend to learn the modality-invariant features of RGB and NI images. So, only conduct metric disruption of $\mathcal{F}_h$ can not get effective attack performance on cross-modality models. To solve this problem, Cross Modality Simulated Disruption method is proposed. CMSD cross-inputs the benign and adversarial images of RGB/NI modalities to NI/RGB modality subnetworks to get the features, which can be formulated by
\begin{equation}
    \mathcal{F}^R_{N}=\mathcal{S}_{R}(x_{N}),\quad \mathcal{F}^{R^\prime}_{N}=\mathcal{S}_{R}(x^{\prime}_{N}),
    \label{eq:eq8}
\end{equation}
\begin{equation}
    \mathcal{F}^N_{R}=\mathcal{S}_{N}(x_{R}),\quad \mathcal{F}^{N^\prime}_{R}=\mathcal{S}_{R}(x^{\prime}_{R}).
    \label{eq:eq9}
\end{equation}
These features can represent the modality-invariant features in cross-modality model embeddings, as they are extracted by non-corresponding cross modality subnetworks without modality-specific feature obtained. CMSD facilitates the transferability of MUA by pushing these cross modality simulated features away by $\mathcal{L}^h_{CMSD}$.
\begin{equation}
    \mathcal{L}^h_{CMSD} =
    \begin{cases}
        -\mathcal{E}(\mathcal{F}^N_{R},\mathcal{F}^{N^\prime}_{R}) & \text{if } h = R \\
        -\mathcal{E}(\mathcal{F}^R_{N},\mathcal{F}^{R^\prime}_{N}) & \text{if } h = N \\
        0                                                              & \text{if } h = T
    \end{cases}
    \label{eq:eq10}
\end{equation}

\subsection{Multi Modality Collaborative Disruption}
\begin{figure}[t]
    \centering
    \includegraphics[width=1.0\columnwidth]{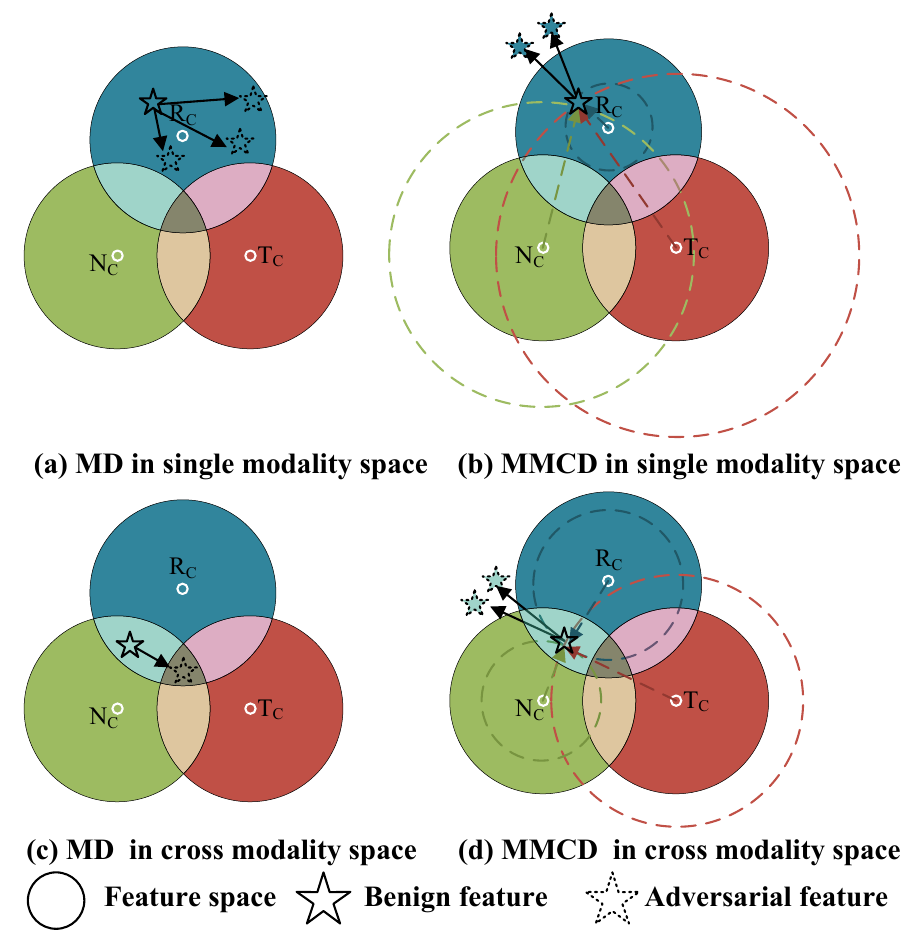} 
    \caption{The comparisons of MD and MMCD. The blue, green and red regions represent distinct feature spaces corresponding to RGB, NI and TI modality images of the same person. The regions where different modality feature spaces intersect are shared feature subspaces across different modalities. $R_c$, $N_c$ and $T_c$ are individual feature centers for each modality. Adversarial features under MD constraints may only move away from the original features and cannot ensure an effective departure away from the individual feature space, whereas under the guidance of MMCD, the adversarial features are constrained to directly far away from the individual feature space.}
    \label{fig:3}
\end{figure}

Simply conducting metric disruption on each modality is inadequate, as it may only compromise the discriminative features, rather than corrupting all features comprehensively. Motivated by the multi-modality re-id models, where the unique features of each modality are leveraged as complementary components for more comprehensive feature fusion to achieve effective recognition, Multi Modality Collaborative Disruption method is invented. MMCD aims to facilitate the complete destruction of person features by the assistance of multi modality auxiliary features. Concretely, our MMCD seeks to push the adversarial feature $\mathcal{F}^\prime$ away from the individual feature centers $\mathcal{F}^c$ of three modalities, which is formulated by 
\begin{equation}
    \mathcal{L}^h_{MMCD} = \sum\nolimits_{m}^{R,N,T}-\mathcal{E}_(\mathcal{F}^\prime_h, \mathcal{F}^c_m).
    \label{eq:eq11}
\end{equation}
The comparison diagram of MMCD and Metric Disruption in Fig. \ref{fig:3} shows the superiority of our MMCD for transferable attack performance on omni-modality re-id models. In contrast to the MD constraints, which merely prompt the adversarial features to deviate from the original features without ensuring a substantial departure from the individual feature space, the MMCD approach deliberately constrains the adversarial features to diverge from the individual feature space by the collaboration of multi modality features.

\subsection{Overall Objective}
Three modality unified adversarial generators $\mathcal{G}_h$ are independently trained with corresponding discriminator $\mathcal{D}_h$ in our paper. The GAN loss of $\mathcal{G}_h$ and $\mathcal{D}_h$ are denoted as
\begin{equation}
    \mathcal{L}^h_\mathcal{D} = \log \mathcal{D}_h(x_h)+\log(1-\mathcal{D}_h(x^{\prime}_{h})),
    \label{eq:eq15}
\end{equation}
\begin{equation}
    \mathcal{L}^h_g = \log(1-\mathcal{D}_h(x^\prime_h)).
    \label{eq:eq12}
\end{equation}
The generated AEs are constrained by the adversarial attack loss, formulated as
\begin{equation}
    \mathcal{L}_{adv}^h = \lambda_1\mathcal{L}^h_{MD}+\lambda_2\mathcal{L}^h_{CMSD} + \lambda_3\mathcal{L}^h_{MMCD},
    \label{eq:eq13}
\end{equation}
where $\lambda_1$, $\lambda_2$ and $\lambda_3$ are the balance weights. The overall objective function on adversarial generator $\mathcal{G}_h$ is formulated by 
\begin{equation}
    \mathcal{L}_{\mathcal{G}}^h = \mathcal{L}_g^h+ \mathcal{L}_{adv}^h.
    \label{eq:eq14}
\end{equation}

\begin{algorithm}[t]
    \small
    \caption{Modality Unified Attack}
    \label{alg:algorithm}
    \renewcommand{\algorithmicrequire}{\textbf{Input:}}
    \renewcommand{\algorithmicensure}{\textbf{Output:}}
    \begin{algorithmic}[1] \color{black}
      \REQUIRE Training data $\mathcal{X}$=\{$x_R$,$x_N$,$x_T$\}, multi-modality agent model $\mathcal{M}_M$, generator $\mathcal{G}_h$ and discriminator $\mathcal{D}_h$ for each modality $h$
      \ENSURE  Modality unified adversarial generator $\mathcal{G}_h$ for each $h$ modal 
      \STATE Initialize parameters $\theta_h$ of $\mathcal{G}_h$, $\varphi_h$ of $\mathcal{D}_h$, learning rate $\alpha$ of them 
      \FOR{$i$=0 to $\mathcal{I}$}
      
      \STATE  Sample batch images $x_h$ and generate AEs $x^{\prime}_{h}$ by Eq.\ref{eq:eq2}
      \STATE  Feed $x_h$ and $x^{\prime}_{h}$ to $\mathcal{M}_M$
      \STATE  Calculate Metric Disruption loss $\mathcal{L}^h_{MD}$ by Eq.\ref{eq:eq7}
      \STATE  Cross-inputs the R/N modalities to N/R modality subnetworks
      \STATE  Calculate Cross Modality Simulated Disruption loss $\mathcal{L}^h_{CMSD}$ by Eq.\ref{eq:eq10}
      \STATE  Compute the feature center of benign multi-modality images and calculate Multi Modality Collaborative Disruption loss $\mathcal{L}^h_{MMCD}$ by Eq.\ref{eq:eq11}
      \STATE Calculate discrimination loss $\mathcal{L}^h_\mathcal{D}$ by Eq.\ref{eq:eq15} and generator loss $\mathcal{L}_{\mathcal{G}}^h$ by Eq.\ref{eq:eq14}
      
      \STATE Update parameters $\theta_h\leftarrow\theta_h-\alpha\nabla_{\theta_h}\mathcal{L}_{\mathcal{G}}^h$
      \STATE Update parameters $\varphi_h\leftarrow\varphi_h-\alpha\nabla_{\varphi_h}\mathcal{L}^h_\mathcal{D}$
      \ENDFOR
    \end{algorithmic}
  \end{algorithm}

\section{Experiments}

\subsection{Experiments Setup}

\textbf{Target models.}
To comprehensively evaluate the attack performance of our MUA on omni-modality re-id models, experiments on numerous re-id models are set up in our paper. Specifically, we take BOT \cite{luo2019bag}, LSRO \cite{zheng2017unlabeled}, MuDeep \cite{qian2017multi}, Aligned \cite{zhang2017alignedreid}, MGN \cite{wang2018learning}, HACNN \cite{li2018harmonious}, Transreid \cite{he2021transreid}, PAT \cite{ni2023part} models as target single-modality models, CAJ \cite{Ye2024Channel, ye2021channel}, DDAG \cite{ye2020dynamic}, PMT \cite{lu2023learning}, MMN \cite{zhang2021towards} models as target cross-modality models, and UniCat \cite{crawford2023unicat}, EDITOR \cite{zhang2024magic} models as target multi-modality models. It is worth noting that these models are representative as there are in different scheme (part-based, global-based, attention-based, data augmentation-based) and different backbones (ResNet \cite{he2016deep}, ViT \cite{dosovitskiy2020image}, DenseNet \cite{huang2017Densely}, Inception-v3 \cite{szegedy2016rethinking} and AGW \cite{ye2021deep}).

\textbf{Target datasets.} For different type of re-id task, we conducted evaluations on classical benchmarks, including the Market \cite{zheng2015scalable}, LLCM \cite{zhang2023diverse}, and RGBNT201 \cite{zheng2021robust} datasets.
Market \cite{zheng2015scalable} is a general single-modality person re-id dataset. It consists of 12,936 training images of 751 persons, 19,732 query images and 3,368 gallery images of 750 persons, captured from 6 cameras. 
The LLCM \cite{zhang2023diverse} dataset features a 9-camera network designed for low-light cross-modality environments, capturing RGB images during the day and infrared images at night. It includes 30,921 training bounding boxes for 713 identities (16,946 RGB and 13,975 infrared) and 13,909 testing bounding boxes for 351 identities (8,680 RGB and 7,166 infrared). Both RGB-to-infrared and infrared-to-RGB modes are used to evaluate cross-modality models.
The RGBNT201 \cite{zheng2021robust} dataset is collected across four non-overlapping views on campus, with triplicated cameras capturing RGB, NI, and TI modalities simultaneously. It consists of 40,000 manually annotated image triplets for 201 identities under diverse illumination and background conditions. For evaluation, 4,787 images per modality are selected, ensuring diverse poses and avoiding redundancy, to support comprehensive experimental analysis.
\begin{table*}[t]
    \centering
    \caption{Comparisons with SOTA methods on omni-modality re-id models. $H^\prime_{1}$-$H_{2}$ means $H_1$ modality images are perturbed by modality unified adversarial generator $G_{H_1}$ to retrieve the benign $H_2$ modality images. The results on aAP and mDR are in \textbf{blod} and the best results are show in \textcolor{blue}{\textbf{blue}}.}
    \begin{threeparttable}
        \begin{tabular}{c|c|c|ccc|cc|c}
            \hline
            \multirow{2}{*}{ Attack Setting } & \multirow{2}{*}{ Models }          & \multirow{2}{*}{ None } & \multicolumn{3}{c|}{ Single-modality Attack } & \multicolumn{2}{c|}{ Cross-modality Attack } & \multicolumn{1}{c}{ Omini-modality Attack }                                                                                  \\
            \cline{4-9}
                                              &                                    &                         & MetaAttack                                    & Mis-Ranking                                  & MUAP                                        & CMPS                & FASM                & \textbf{Ours}                    \\
            \hline
            \multirow{10}{*}{\textbf{\shortstack{$\mathcal{M}_{S}$                                                                                                                                                                                                                                                                         \\ ($R^\prime$-$R$)}}}
                                              & BOT                                & 85.4                    & 26.3                                          & 46.3                                         & 42.9                                        &  72.3                    & 57.6                     & 29.4                             \\
                                              & LSRO                               & 77.2                    & 68.6                                          & 36.7                                         & 35.7                                        &  66.9                    &   47.9                   & 29.5                             \\
                                              & MuDeep                             & 49.9                    & 37.8                                          & 11.9                                         & 9.7                                         & 39.9                     & 23.2                     & 10.0                             \\
                                              & Aligned                            & 79.1                    & 59.4                                          & 47.5                                         & 48.0                                        &  66.0                    &  54.5                    & 37.6                             \\
                                              & MGN                                & 82.1                    & 73.0                                          & 46.7                                         & 40.6                                        & 72.8                     &  55.2                    & 39.1                             \\
                                              & HACNN                              & 75.2                    & 63.9                                          & 27.0                                         & 23.8                                        &70.3                      &  52.3                    & 19.8                             \\
                                              & Transreid                          & 86.6                    & 80.0                                          & 65.2                                         & 58.3                                        &  81.9                    & 71.8                     & 58.0                             \\
                                             
                                              & PAT                                & 78.4                    & 67.7                                          & 63.4                                         & 59.7                                        &   70.0                   & 60.6                    & 47.3                             \\
                                              
                                              &  \textbf{aAP$\downarrow$}&  \textbf{76.7} & \textbf{59.6}                      & \textbf{43.1}                      & \textbf{39.8}                    &  \textbf{67.5}                     &   \textbf{52.9}                   & \textcolor{blue}{\textbf{33.8}}\\
                                              & \textbf{mDR$\uparrow$ }  & \textbf{0.0}  & \textbf{22.3}                     & \textbf{43.8}                      & \textbf{48.1}                    &  \textbf{12.0}                   &   \textbf{31.1}                  & \textcolor{blue}{\textbf{55.9}} \\
            \hline
            \multirow{7}{*}{\textbf{\shortstack{ $\mathcal{M}_{C}$                                                                                                                                                                                                                                                                         \\ ($R^\prime$-$N$)}}}
                                              & CAJ                                & 59.8                    & 55.9                                          & 56.5                                         & 52.0                                        & 52.8                     & 55.1                     & 45.0                             \\
                                              & DDAG                               & 53.0                    & 46.5                                          & 45.8                                         & 41.9                                        &  28.5                    &  33.0                    & 38.3                             \\
                                              & MMN                                & 62.7                    & 54.8                                          & 54.6                                         & 50.6                                        &   37.9                   &    44.1                  & 46.3                             \\
                                              & PMT                                & 65.0                    & 59.0                                          & 58.6                                         & 58.0                                        & 48.7                     & 54.1                     & 52.2                             \\
                                              & \textbf{aAP$\downarrow$} & \textbf{60.1}& \textbf{54.1}                    & \textbf{53.9}                    & \textbf{50.6}                  &  \textcolor{blue}{\textbf{42.0}}                &     \textbf{46.6}                  & \textbf{45.5} \\
                                              & \textbf{mDR$\uparrow$ }  & \textbf{0.0}  & \textbf{10.1}                    & \textbf{10.4}                     & \textbf{15.8}                   &   \textcolor{blue}{\textbf{30.1}}                     &   \textbf{22.5}            & \textbf{24.4} \\

            \hline
            \multirow{7}{*}{\textbf{\shortstack{$\mathcal{M}_{C}$                                                                                                                                                                                                                                                                          \\ ($N^\prime$-$R$)}}}
                                              & CAJ                                & 56.6                    & -                                             & -                                            & -                                           & 39.4                     & 38.6                     & 29.0                             \\
                                              & DDAG                               & 49.6                    & -                                             & -                                            & -                                           &  19.0                   &   15.7                   & 24.4                             \\
                                              & MMN                                & 58.9                    & -                                             & -                                            & -                                           &  23.7                    &   21.4                   & 28.4                             \\
                                              & PMT                                & 60.6                    & -                                             & -                                            & -                                           &   28.7                   &   30.7                   & 33.3                             \\
                                              & \textbf{aAP$\downarrow$ }& \textbf{56.1} & -                      & -                       & -                       &   \textbf{27.7}                  &    \textcolor{blue}{\textbf{26.6}}                 &\textbf{28.8} \\
                                              & \textbf{mDR$\uparrow$ }  & \textbf{0.0} & -                         & -                       & -                      &    \textbf{51.0}                &   \textcolor{blue}{\textbf{52.8}}                   & \textbf{49.0 }\\
            \hline
            \multirow{4}{*}{\textbf{\shortstack{$\mathcal{M}_{M}$                                                                                                                                                                                                                                                                          \\ ($R^{\prime}N^{\prime}T^{\prime}$-$RNT$)}}}
                                              & EDITOR                             & 65.7                    & -                                             & -                                            & -                                           & -                    & -                    & 19.2                             \\
                                              & UniCat                             & 57.0                    & -                                             & -                                            & -                                           & -                    & -                    & 26.6                             \\
                                              & \textbf{aAP$\downarrow$} & \textbf{61.4} & -                         &  -                       &  -                        &  - &  - & \textcolor{blue}{\textbf{22.9}} \\
                                              & \textbf{mDR$\uparrow$}   & \textbf{0.0} & -                         & -                      & -                     & -& -& \textcolor{blue}{\textbf{62.7}}\\
            \hline
        \end{tabular}
        \begin{tablenotes}
            \item The symbol `-' means that corresponding attack methods can not generate perturbations on specific attack settings.
        \end{tablenotes}
    \end{threeparttable}
    \label{tab:1}
\end{table*}

\textbf{Evaluation metrics.} Mean Average Precision (mAP) \cite{zheng2015scalable}, average mAP (aAP) and mean mAP Drop Rate (mDR) \cite{ding2021beyond} are used to evaluate the adversarial attack performance of the generated AEs against the different omni-modality re-id models. The mAP measures the retrieval results of single model, aAP is utilized to assess the average mAP across multiple models by
\begin{equation}
  aAP =\ \frac{\sum_{i=0}^{N}{mAP}_i}{N},
\end{equation}
where ${mAP}_i$ represents mAP of the $i$-th re-id models. The mDR is adopted to show the success rate of the adversarial attacks to multiple re-id models by calculating the drop rate of $aAP$ after attacking by
\begin{equation}
  mDR =\frac{aAP-aAP^{\prime}}{aAP},
\end{equation}
where $aAP^{\prime}$ is evaluated on generated adversarial examples.

\textbf{Implementation details.} 
The multi-modality re-id model TOP \cite{wang2024top} trained on RGBNT201 \cite{zheng2021robust} benchmark is treated as surrogate model in our MUA. The adversarial generator and discriminator are referenced to Mis-Ranking \cite{wang2020transferable}. The adversarial generator $\mathcal{G}_N$ for NI images outputs one channel perturbations and others for three channel perturbations, considering different imaging mechanisms and the demand for imperceptible perturbations. All experiments are performed by $\mathcal{L}_{\infty}$-bounded attacks with $\epsilon=8/255$, where $\epsilon$ is the upper bound for the change of each pixel. Adam \cite{kingma2014adam} optimizer is employed to optimize the model parameters with a learning rate of 2e-4. The balance weights $\lambda_1$, $\lambda_2$ and $\lambda_3$ are set to 50, 50, and 10, respectively.

\subsection{Comparisons with State-of-the-art Methods}

We compare our proposed MUA method with state-of-the-art (SOTA) attack methods on black-box re-id models,
including MUAP \cite{ding2021beyond}, Mis-Ranking \cite{wang2020transferable}, MetaAttack \cite{yang2022towards}, CMPS \cite{gong2024cross}, FSAM \cite{yangfeature}. It is worth noting that MUAP \cite{ding2021beyond}, Mis-Ranking \cite{wang2020transferable}, MetaAttack \cite{yang2022towards} methods are designed only for producing perturbations on RGB images to attack single-modality models and CMPS \cite{gong2024cross}, FSAM \cite{yangfeature} are designed for producing perturbations on RGB and NI images to attack cross-modality models. Considering that our surrogate model and training data differ from the target single- and cross-modality models and their corresponding training data, leading to cross-dataset and cross-model black-box attack scenarios, we compare the performance of these SOTA methods in cross-dataset and cross-model settings on the target single-modality and cross-modality models for a fair evaluation. However, for the multi-modality re-ID task, there is only one RGBNT201 \cite{zheng2021robust} dataset available, so we focus solely on testing the cross-model transferability. Specifically, for single-modality attack methods, we retrained the attack methods on the IDE\cite{zheng2016person} as the surrogate model and used DukeMTMC\cite{ristani2016performance} as the training data, testing on the target model trained on the Market dataset. For cross-modality, we trained on the  AGW \cite{ye2021deep} model and SYSU-MM01 \cite{wu2017rgb} dataset, and tested on the target model trained on the LLCM \cite{zhang2023diverse} dataset. Since the CMPS \cite{gong2024cross} method generates a universal three-channel perturbation for RGB and NI images, which is unreasonable because applying a three-channel perturbation to a single-channel NI image violates the imaging principles of NI images, we average the three-channel perturbation generated by this method into a single channel and add it to the NI image when testing NI-to-RGB cross-modal adversarial retrieval.  For multi-modality re-ID models, we trained on the surrogate model TOP \cite{wang2024top} and tested on the target models UniCat \cite{crawford2023unicat} and EDITOR \cite{zhang2024magic}.

\textbf{Comparisons on single-modality models.} It can be seen in Tab. \ref{tab:1} that our method are superior to SOTA methods in attacking single-modality models $\mathcal{M}_{S}$, even though these SOTA methods are specifically designed for single-modality models and ours is for the omni-modality re-id models, which powerfully illustrates the advantages of our proposed method. Specifically, our methods our outperforms the SOTA methods by 6.0\% and 7.8\% on aAP and mDR, respectively.
\begin{table*}[t]
    \centering
    \caption{Performance analysis of each component in our MUA. $H^\prime_{1}$-$H_{2}$ means $H_1$ modality images are perturbed by modality unified adversarial generator $G_{H_1}$ to retrieve the benign $H_2$ modality images. MMCD and MMCD$^\prime$ represent multi modality collaborative disruption on original features extracted from surrogate model in MD and simulated cross modality features in CMSD, respectively. The results on aAP and mDR are in \textbf{blod} and the best results are show in \textcolor{blue}{\textbf{blue}}.}
    \begin{tabular}{c|c|c|c|c|c|c}
        \hline
        Attack setting & Models                    & None           & +MD            & +MMCD          & +CMSD          &+MMCD$^\prime$         \\
        \hline
        \multirow{10}{*}{\textbf{\shortstack{$\mathcal{M}_{S}$                                                                                                                            \\ ($R^\prime$-$R$)}}}
                   & BOT                                & 85.4                    & 51.2                    & 40.9                    & 50.4                    & 29.4                    \\
                   & LSRO                               & 77.2                    & 36.3                    & 30.7                    & 39.9                    & 29.5                    \\
                   & MuDeep                                & 49.9                   & 8.8                    & 7.4                    & 11.7                    & 10.0                    \\
                   & Aligned                            & 79.1                    & 43.1                    & 37.5                    & 42.6                    & 37.6                    \\
                & MGN                                & 82.1                    & 42.7                    & 37.0                    &46.9                    & 39.1                    \\
                   & HACNN                              & 75.2                    & 26.1                    & 21.4                    & 28.1                    & 19.8                    \\
                   & Transreid                          & 86.6                    & 54.3                    & 50.4                    & 54.3                    & 58.0                    \\
                   & PAT                                & 78.4                    & 50.4                    & 50.1                    & 44.1                    & 47.3                    \\
                   & \textbf{aAP$\downarrow$} &  \textbf{76.7} &  \textbf{39.1} & \textbf{34.4} & \textbf{39.8} & \textcolor{blue}{\textbf{33.8}}\\
                   & \textbf{mDR$\uparrow$}   & \textbf{0.0}  & \textbf{49.0} & \textbf{55.1} & \textbf{48.2} & \textcolor{blue}{\textbf{55.9}} \\
        \hline
        \multirow{7}{*}{\textbf{\shortstack{ $\mathcal{M}_{C}$                                                                                                                            \\ ($R^\prime$-$N$)}}}
                 
                   & CAJ                              & 59.8                    & 55.9                    & 56.5                    & 47.0                    & 45.0                    \\
                   & DDAG                               & 53.0                    & 46.5                    & 45.8                    & 36.9                    & 38.3                    \\
                   & MMN                                & 62.7                    & 54.8                    & 54.6                    & 45.6                    & 46.3                    \\
                   & PMT                                & 65.0                    & 59.0                    & 58.6                    & 53.0                    & 52.2                    \\
                   & \textbf{aAP$\downarrow$} & \textbf{60.1} & \textbf{54.1} & \textbf{53.9} &  \textbf{45.6} & \textcolor{blue}{\textbf{45.5}}\\
                   & \textbf{mDR$\uparrow$}   & \textbf{0.0 } & \textbf{10.1}  & \textbf{10.4}  &  \textbf{24.1 }& \textcolor{blue}{\textbf{24.4}} \\
        \hline
        \multirow{7}{*}{\textbf{\shortstack{$\mathcal{M}_{C}$                                                                                                                             \\ ($N^\prime$-$R$)}}}
                 
                   & CAJ                                & 56.6                    & 34.7                    & 33.7                    & 30.1                    & 29.0                    \\
                   & DDAG                               & 49.6                    & 27.5                    & 27.8                    & 25.7                    & 24.4                    \\
                   & MMN                                & 58.9                    & 32.9                    & 28.6                    & 28.4                    & 28.4                    \\
                   & PMT                                & 60.6                    & 41.2                    & 42.2                    & 36.7                    & 33.3                    \\
                   & \textbf{aAP$\downarrow$} & \textbf{56.4} & \textbf{34.1} & \textbf{33.1} & \textbf{30.2} & \textcolor{blue}{\textbf{28.8}} \\
                   & \textbf{mDR$\uparrow$ }  & \textbf{0.0  }& \textbf{39.6} & \textbf{41.4} & \textbf{46.4} &\textcolor{blue}{\textbf{49.0}} \\
        \hline
        \multirow{4}{*}{\textbf{\shortstack{$\mathcal{M}_{M}$                                                                                                                             \\ ($R^{\prime}N^{\prime}T^{\prime}$-$RNT$)}}}
                   & EDITOR                                & 65.7                    & 22.8                    & 21.1                    & 21.4                    & 19.2                    \\
                   & UniCat                             & 57.0                   & 28.0                    & 28.3                    & 33.8                    & 26.6                    \\
                   & \textbf{aAP$\downarrow$} & \textbf{61.4} & \textbf{25.4} & \textbf{24.7} & \textbf{27.6} & \textcolor{blue}{\textbf{22.9}} \\
                   & \textbf{mDR$\uparrow$}   & \textbf{0.0}  & \textbf{58.6} & \textbf{59.8 }& \textbf{55.0} & \textcolor{blue}{\textbf{62.7}} \\
        \hline
    \end{tabular}
  
    \label{tab:2}
\end{table*}

\textbf{Comparisons on cross-modality models.} The performance of cross-modality models $\mathcal{M}_{C}$ are evaluated by RGB to NI and NI to RGB retrieval modes. For RGB to NI mode, the RGB AEs generated by our method demonstrate transferability comparable to SOTA cross-modality attack methods specifically designed for this task, achieving 45.5\% in aAP and 24.4\% in mDR metrics. For NI to RGB mode, our approach also achieves an effective attack effect with 28.1\% aAP and 49.0\% mDR, which only marginally lower than the SOTA dedicated methods by 2.2\% and 3.8\% in aAP and mDR, respectively.

\textbf{Comparisons on multi-modality models.} For multi-modality re-id models, our method is the only attack that generates AEs for RGB, NI and TI images to disrupt their performance. Results in Tab. \ref{tab:1} show that our method can effectively attack multi-modality re-id models with 22.9\% aAP and 62.7\% mean mAP drop rate. 

\subsection{Ablation Studies}
The ablation study results of MD, MMCD, CMSD and MMCD$^\prime$ are presented in Tab. \ref{tab:2}. MMCD and MMCD$^\prime$ represent multi modality collaborative disruption on original features extracted from surrogate model in MD and simulated cross modality features in CMSD, respectively.

\textbf{Effectiveness of MD.} Tab. \ref{tab:2} shows that all three types of models have some degree of performance degradation after MD constraint. Specifically, the application of MD gets 49.0\%, 10.1\%, 39.6\% and 58.2\% mDR on omni-modality models. This indicates that our selection of the multi-modality model as the surrogate model and destruction of the intermediate layer features before fusion is effective.

\textbf{Effectiveness of MMCD.} \textcolor{black}{MMCD is conducted on adversarial intermediate layer features in MD.} It can be observed from Tab. \ref{tab:2} that the incorporation of MMCD results in increases of 6.1\%, 0.3\%, 1.8\% and 1.2\% in mDR on omni-modality models, which proves the effectiveness of proposed MMCD.

\textbf{Effectiveness of CMSD.} As can be seen in Tab. \ref{tab:2}, under the constraints of MD and MMCD, the mean mAP Drop Rate of the cross-modality models is still modest at 10.4\% and 41.4\%. When CMSD is added into the training, the mDR increases from 10.4\% to 24.1\% in RGB retrieves NI mode and from 41.4\% to 46.4\% in NI retrieves RGB mode, which demonstrates our CMSD's effectiveness on cross-modality re-id models.


\textbf{Effectiveness of MMCD$^\prime$.} Although the performances of cross-modality models drop a lot after incorporating the CMSD, the attack performances against single-modality and multi-modality models degrade due to the additional different feature embedding constraints. MMCD$^\prime$ is conducted on simulated cross modality features in CMSD to further facilitate the adversarial corruptions of simulated cross modality features. 
Tab. \ref{tab:2} shows the advantage of MMCD$^\prime$, where the mDR of single-modality and multi-modality models increases by 7.7\% and 7.7\%, respectively. For cross-modality models, the mDR also increases by 2.6\% in NI retrieves RGB mode. It is worth noting that all the attack performances get better after CMSD and MMCD guidance, proving the effectiveness of our approach on cross modality feature simulation and multi modality collaborative disruption on simulated features. 
\begin{figure}[t]
    \centering
    \includegraphics[width=1.0\columnwidth]{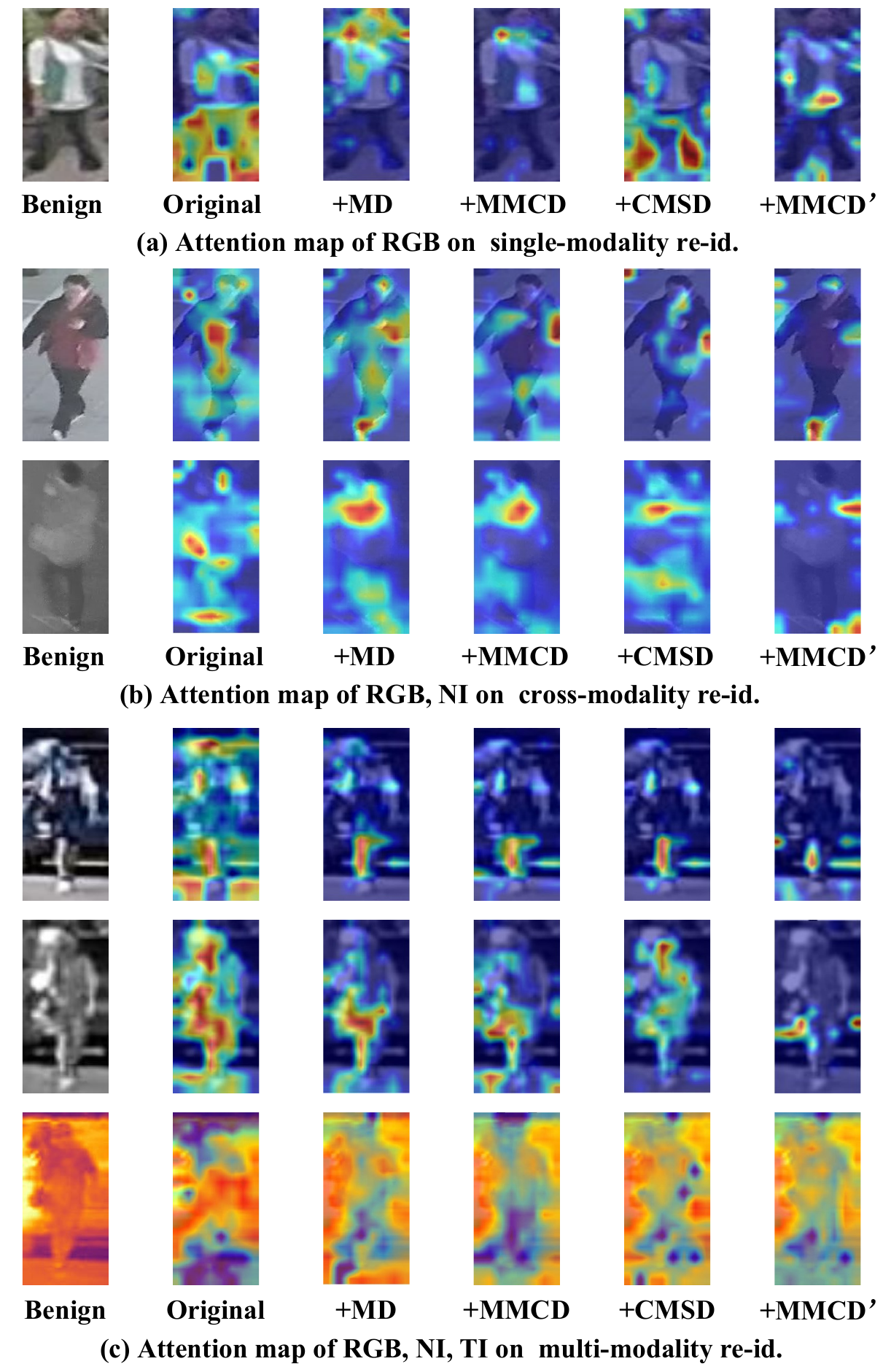}
    \caption{\textcolor{black}{Attention maps of RGB, NI, and TI modalities for both benign and adversarial examples across different re-ID models. The original attention maps correspond to benign images, while the others visualize adversarial examples generated under different loss constraints (MD, MMCD, CMSD). Transreid \cite{he2021transreid}, MMN \cite{zhang2021towards} and EDITOR \cite{zhang2024magic} are token as the single-, cross- and multi-modality re-id models for Visualizations.}}
    \label{fig:5}
\end{figure}

    \textcolor{black}{To further validate the effectiveness of our proposed methods, we visualize the attention maps for both benign and adversarial examples of RGB, NI, and TI modalities on different types of re-id models. The results in Fig. \ref{fig:5}(a) show the attention maps of RGB on single-modality re-id model, which illustrates the MMCD and MMCD$^\prime$ can eliminate the attention and CMSD guided different attention attack on the same images. For attention maps of cross-modality re-id model in Fig. \ref{fig:5}(b), the MMCD and CMSD continuously disrupt the attentions of target person on both RGB and NI images. There are also obviously comprehensive disruptions of three modality images on multi-modality re-id with our methods, shown in Fig. \ref{fig:5}(c). In summary, the visualization results demonstrate that our method effectively and comprehensively perturbs feature representations across different re-id tasks (single-modality, cross-modality, and multi-modality) and modalities (RGB, NI, TI), thereby validating its strong adversarial transferability and attack efficacy.}

\begin{figure}[t]
    \centering
    \includegraphics[width=1.0\columnwidth]{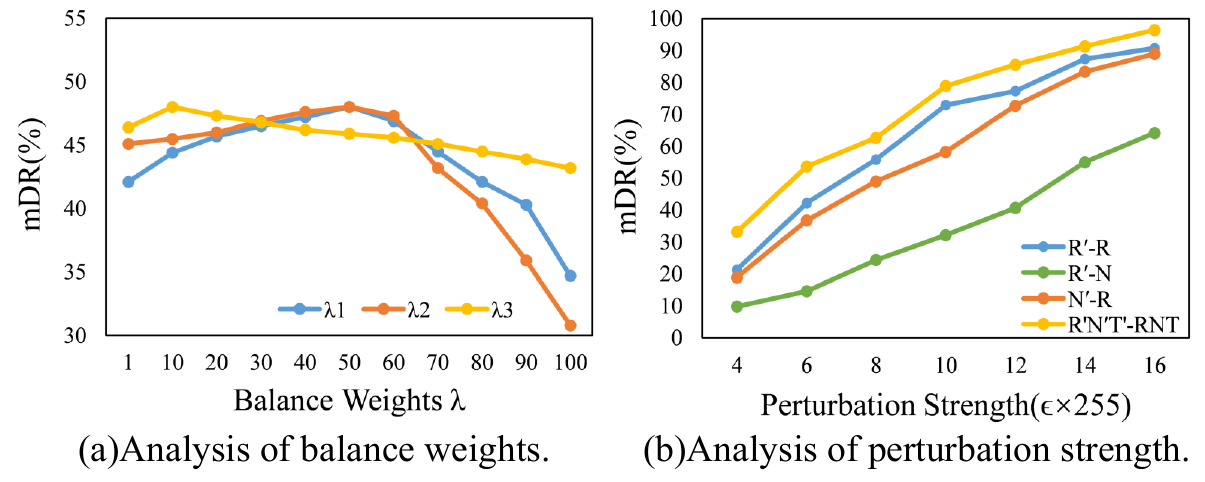}
    \caption{\textcolor{black}{Analysis of mDR under different perturbation strength and balance weights. The mDR values in the balance weights analysis represent the average across four retrieval settings.}}
    \label{fig:6}
\end{figure}

\begin{table}[t]
    \centering
    \caption{\textcolor{black}{Attack effectiveness against defense methods.}}
    \resizebox{0.5\textwidth}{!}{
    \begin{tabular}{c|c|cc|c|c}
        \hline Re-id models & Method & Randomization & JPEG & $aAP$ & $mDR$ \\
        \hline Single-modality & None & 84.6 & 83.8 & 84.2 & - \\
        ($R^{\prime}$-$R$) & MUA & 45.2 & 54.1 &49.7 &41.0 \\
        \hline Cross-modality & None & 59.6 & 58.4 & 59.0 & - \\
        ($R^{\prime}$-$N$) & $M U A$ & 49.2 & 47.4 & 48.3 & 18.1 \\
        \hline Cross-modality & None & 54.3 & 52.2 & 53.3 & - \\
        ($N^{\prime}$-$R$) & $M U A$ & 32.8 & 31.7 & 32.3 & 39.5 \\
        \hline Multi-modality & None & 60.4 & 59.5 & 60.0 & - \\
        ($R^{\prime}N^{\prime}T^{\prime}$-$RNT$) & $M U A$ & 39.1 & 36.9 & 38.0 & 36.7 \\
        \hline
\end{tabular}
\label{tab:defense}
    }
\end{table}

    \textcolor{black}{\textbf{Analysis of key parameters.} We conduct analysis on the balance weights $\lambda_1$, $\lambda_2$ and $\lambda_3$ and different perturbation strength $\epsilon$. To balance the influence of different loss terms in Eq. \ref{eq:eq13}, we employ these $\lambda_1$, $\lambda_2$, $\lambda_3$ three weighting coefficients. We investigate their individual impacts through an ablation study, where we systematically vary one hyperparameter while keeping the others fixed.  Fig. \ref{fig:6}(a) illustrates the impact of balance weights, showing that high $\lambda_1$ and $\lambda_2$ values lead to notable mDR declines, potentially forcing overfitting to surrogate model feature disruption. In contrast, performance remains stable across varying $\lambda_3$ values, demonstrating the stabilizing effect of the MMCD loss. On top of the experiments, we set $\lambda_1$=50, $\lambda_2$=50,
    and $\lambda_3$=10, which achieves the best performance. For perturbation strength, larger values improve attack performance, but we chose 8/255 to balance imperceptible perturbations.}

\subsection{Attack Effectiveness against Defense Method}

\textcolor{black}{Tab. \ref{tab:defense} provides supplementary experimental results analyzing the robustness of our approach when confronted with different defense mechanisms. These include input transformation defenses such as JPEG compression \cite{das2017keeping} with a quality factor set to 60\%, as well as noise-based mitigation approaches like Randomization \cite{xie2018mitigating}. The experiments were conducted using the single-modality BOT \cite{luo2019bag} model, cross-modality MMN \cite{zhang2021towards} and multi-modality EDITOR \cite{zhang2024magic} model as the target system for evaluation. 
It is notable that our method also demonstrates attack effectiveness against these defenses, with 41.0\%, 18.1\%, 39.5\% and 36.7\% mDR for omni-modality re-id models.}

\begin{figure*}[t]
    \centering
    \includegraphics[width=2.0\columnwidth]{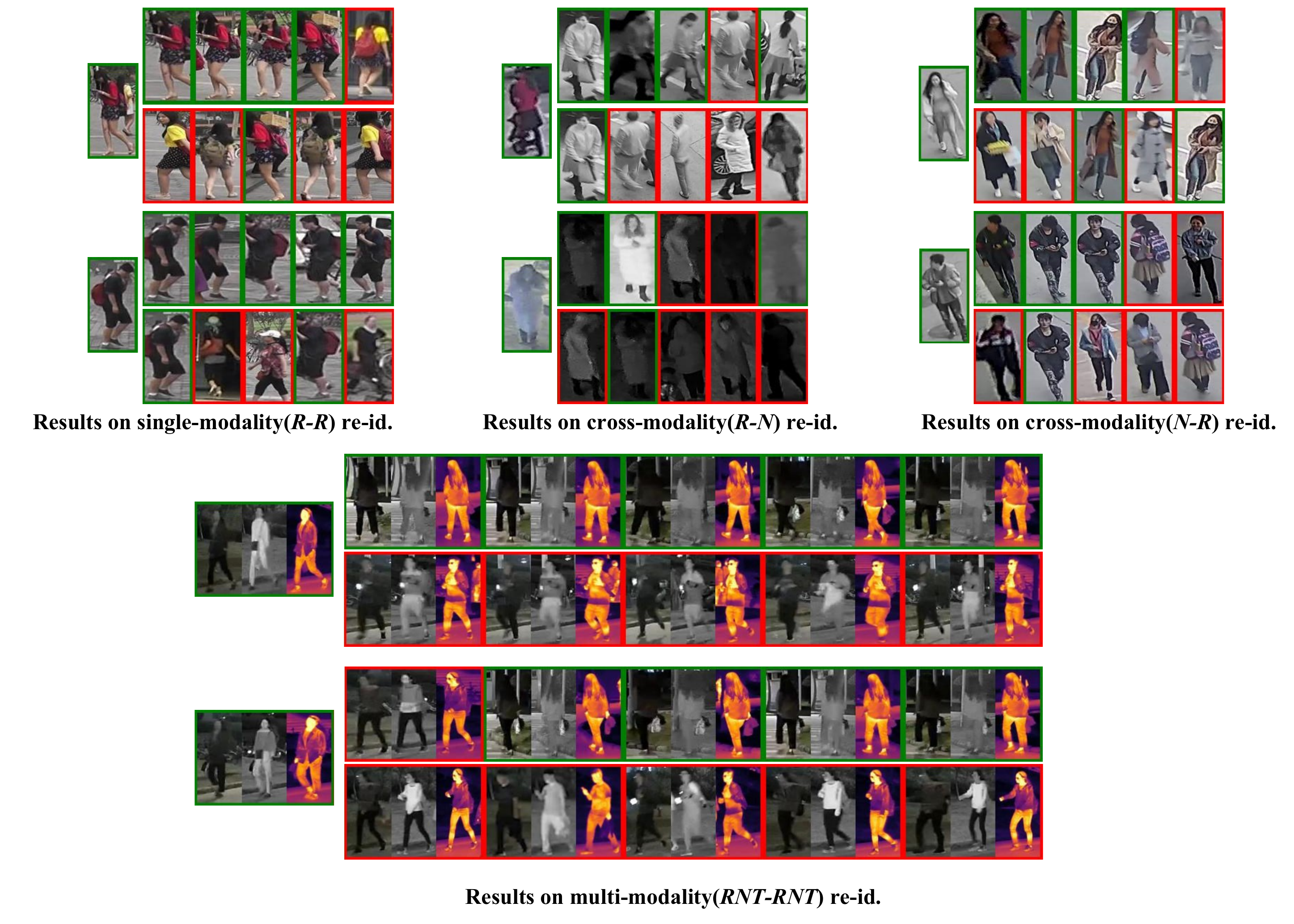}
    \caption{\textcolor{black}{The rank-5 retrieval results of single-. cross- and multi-modality models before and after our attack. Green boxes denote correctly matched images,
    red boxes indicate mismatched images, and the first row  represents the benign results and the second row represents the adversarial results.}}
    \label{fig:4}
\end{figure*}

\subsection{Visualization}

\begin{figure}[t]
    \centering
    \includegraphics[width=1.0\columnwidth]{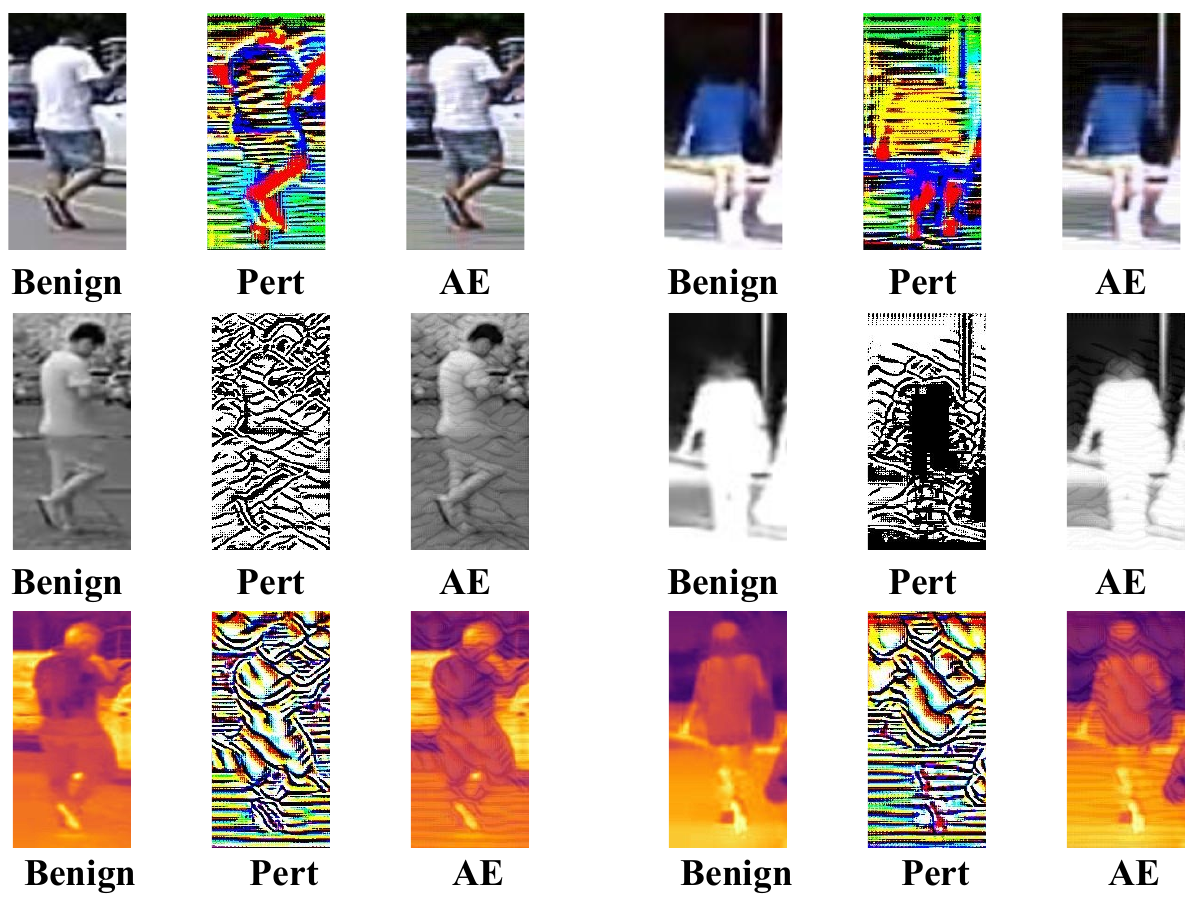}
    \caption{\textcolor{black}{Visualization of perturbations (Pert) and adversarial examples (AE) generated by our MUA framework across three modalities: RGB (top), NI (middle), and TI (bottom).}}
    \label{fig:7}

\end{figure}

    \textcolor{black}{We visualize the adversarial examples and perturbations generate by our MUA across different modality images. As Fig. \ref{fig:7} shows, the perturbations on RGB images exhibit semantically meaningful patterns correlated with target pedestrians and texture perturbations, while those on NI and TI modalities primarily focus only on texture perturbations. This observation aligns well with our expectations for perturbations on different modalities: RGB images are more susceptible to semantic-level perturbations due to their rich visual information and texture perturbations can also attack them in cross-modality tasks, whereas NI and TI modalities, lacking explicit semantic features, achieve better adversarial transferability through texture-based perturbations that preserve modality characteristics while effectively deceiving models. Furthermore, the adversarial examples across all modalities demonstrate imperceptible perturbations. Our modality-specific generators effectively produce perturbations that are carefully adapted to each modality's unique imaging characteristics.}
    
    \textcolor{black}{We also provide visualization of attack results by showcasing the Top-5 retrieval results from the different types target re-id models (i.e., Transreid \cite{he2021transreid} (single-modality), MMN \cite{zhang2021towards} (cross-modality) and EDITOR \cite{zhang2024magic} (multi-modality)). Fig. \ref{fig:4} demonstrates that our method successfully launches effective attacks against various model architectures, significantly compromising their retrieval performance.}

\section{Discussion}
\textcolor{black}{\textbf{Impact.} The proposed adversarial attack method highlights potential security risks in surveillance systems by demonstrating effective attacks against different types of person re-identification models. These findings expose vulnerabilities that could be exploited by malicious actors, emphasizing the need for more robust defenses. At the same time, the method provides a valuable benchmark for evaluating and improving model robustness. In future work, we will explore how the generated adversarial examples can be used to develop more resilient re-identification systems through adversarial training and robustness testing.}

\textcolor{black}{\textbf{Limitation and future work.} The proposed MUA method employs three modality-specific adversarial generators to enable comprehensive omni-modality re-id attacks. While effective, this multi-generator architecture introduces implementation complexity and parameter redundancy. To address these limitations, our future work will develop a diffusion-based unified framework that simultaneously generates effective adversarial samples across all modalities and preserves the distinctive imaging characteristics of each modality.}

\section{Conclusion}


In this paper, we propose the first modality unified attack for omni-modality (single-, cross-, multi-modality) re-id models. Cross Modality Simulated Disruption method is introduced to attack the shared features of the same modality in omni-modality models by simulating features embeddings across different types of re-id models. Multi Modality Collaborative Disruption approach is designed for the attacker to use multi modality features to collaboratively measure the feature disruption and comprehensively damage the information content of perdestrain images. Extensive experiments on various single-, cross- and multi-modality black-box models show that our method can effectively attack the omni-modality re-id models.



\bibliography{egbib}{}
\bibliographystyle{IEEEtran}

\newpage

\end{document}